%% file: acl_latex.tex
\pdfoutput=1

\documentclass[11pt]{article}

\usepackage{acl}

\usepackage{times}
\usepackage{latexsym}
\usepackage{multicol, multirow}

\usepackage[T1]{fontenc}

\usepackage[utf8]{inputenc}

\usepackage{microtype}

\usepackage[disable]{todonotes}

\usepackage{xcolor,colortbl}
\definecolor{Gray}{gray}{0.85}

\newcommand\Tstrut{\rule{0pt}{0.9ex}}       %
\newcommand\Bstrut{\rule[-0.9ex]{0pt}{0pt}} %
\newcommand{\TBstrut}{\Tstrut\Bstrut} %
\definecolor{mintbg}{rgb}{.63,.79,.95}

\title{\textit{Read Top News First}: A Document Reordering Approach for Multi-Document News Summarization}

\author{Chao Zhao$^{1}$\thanks{\quad Equal Contribution} \qquad Tenghao Huang$^{1}$\footnotemark[1] \qquad Somnath Basu Roy Chowdhury$^{1}$ \\ \bf
Muthu Kumar Chandrasekaran \qquad Kathleen McKeown$^{2}$ \qquad Snigdha Chaturvedi$^{1}$ \\
\texttt{\{zhaochao, tenghao, somnath, snigdha\}@cs.unc.edu}\\\texttt{cmkumar087@gmail.com, kathy@cs.columbia.edu}\\$^{1}$ UNC Chapel Hill \qquad $^{2}$ Columbia University}

\usepackage{booktabs}
\usepackage{mathtools}
\usepackage{amsfonts}

\usepackage{amsmath}
\usepackage{hyperref}

\usepackage{xcolor, soul}
\sethlcolor{lime}

\begin{document}

\maketitle
\begin{abstract}
A common method for extractive multi-document news summarization is to re-formulate it as a single-document summarization problem by concatenating all documents as a single meta-document. However, this method neglects the relative importance of documents. We  propose a simple approach to reorder the documents according to their relative importance before concatenating and summarizing them. The reordering makes the salient content easier to learn by the summarization model.
Experiments show that our approach outperforms previous state-of-the-art methods with more complex architectures. 
\end{abstract}

\section{Introduction}
Multi-document news extractive summarization (MDS) aims to extract the salient information from multiple related news documents into a concise summary.
Some approaches use task-specific architectures for this problem.
For example, \citet{wang2020heterogeneous} organize multiple documents as a heterogeneous graph before summarizing them. \citet{zhong2020extractive} formulate the extractive summarization
task as a semantic matching problem. 
Recent works also explored reformulating this problem as a single-document summarization (SDS) problem by concatenating all documents into a single \textit{meta-document} and then using an SDS model to summarize it \cite{cao2017improving,liu2018generating,lebanoff2018adapting,fabbri2019multi}. 

Due to the conventions of news writing~\cite{hong2014improving,hicks2016writing}, salient %
information often appears at the beginning of a news article. As a result, many  summarization systems, including recent %
neural models \cite{kedzie2018content, zhong-etal-2019-searching}, %
{pay more attention to the beginning of the document.} %
{Therefore, in MDS, it is important to consider the order in which the documents are concatenated to form the \textit{meta-document} before applying the summarization model. }

Specifically, we argue that the various documents in the input are not equally important. Some documents contain more salient or detailed information and are more important. %
Therefore, compared with concatenating documents in an arbitrary order, it would be beneficial to reorder the documents such that the important ones are in the front of the meta-document and it becomes easier for the summarization model to learn the salient content.  %

Motivated by these factors, we propose a simple yet effective approach to reorder the input documents according to their relative importance %
before applying a summarization model. We evaluate the effectiveness of our approach on Multi-News \cite{fabbri2019multi} and DUC-2004. \footnote{https://duc.nist.gov/data.html} 
Results show that our simple reordering 
approach significantly outperforms the state-of-the-art methods with more complex model architectures. We also observe that this approach brings more performance gain with the increase in the number of input documents.

\section{Method}
\label{sec::method}

We refer to $ \mathcal{D}$ as a meta-document of $m$ documents $\{d_1, \ldots, d_m\}$ %
with $n$ sentences $\left\{s_1,...,s_n\right\}$ in total. The goal in extractive summarization is to extract a subset of sentences in $\mathcal{D}$ to summarize the input documents. It is usually formulated as a binary sentence classification problem, where each sentence is assigned  a $\{0,1\}$ label to determine if it is to be included in the summary. 

Below, we introduce our document reordering approach, and then {the base} %
summarization model. 

\subsection{Document Reordering}
Document reordering aims to rearrange documents of the meta-document in order of their salience. It can be formulated as determining the relative importance score of each document and then reordering the documents according to their importance scores. %
Here we propose a supervised approach and an unsupervised approach for this task.

\noindent \textbf{Supervised Approach.} {In this approach, we use a BERT \cite{devlin2019bert} based model to learn document importance scores. For this, we first concatenate the documents together while inserting a \texttt{[CLS]} and a \texttt{[SEP]} token at the start and the end of each document. We then encode the concatenated documents using BERT } %
to get the document representation $t_i\in \mathbb{R}^{K}$, which is the representation of the \texttt{[CLS]} token preceding it. To enhance the model's ability to capture the inter-document relationships, we use a 2-layer Transformer to encode $t_i$ and finally obtain a document's \textit{contextualized} representation $h_i \in \mathbb{R}^{K}$.
\begin{equation}
\begin{aligned}
    t_1, \ldots, t_m &= \text{BERT}(d_1,\ldots,d_m)\\
    h_1, \ldots, h_m &= \text{Transformer}(t_1,\ldots,t_m)
\end{aligned}
\end{equation}

Thereafter, in order to predict the importance score for the $i$-th document, $\hat{y}_i$, we apply a linear transformation with a Softmax function.
\begin{equation}
\hat{y}_i = \text{softmax} \left(W h_{i}+b\right),
\label{eq::doc_rank}
\end{equation}
where $W\in \mathbb{R}^{K \times K}$ and $b\in \mathbb{R}^K$ are parameters. 

During training, 
we determine the oracle importance score of each document $d_i$ as the normalized ROUGE-1 F score \footnote{We also tried ROUGE-2 F or ROUGE-1 R but didn't observe a significant difference.} between $d_i$ and the gold abstractive summary $S$:
\begin{equation}
y_i = \frac{\text{ROUGE}(d_i, S)}{\sum_i \text{ROUGE}(d_i, S)}.
\end{equation}

Our learning objective is to minimize the Kullback–Leibler divergence between the predicted distribution $\hat{y}=\{\hat{y}_1, \ldots, \hat{y}_m\}$ and the oracle distribution $y=\{y_1, \ldots, y_m\}$ of importance scores.
\begin{equation}
\mathcal{L} = \text{KL}(\hat{y}, y)
\end{equation}

We train the document reordering model on the training set based on this learning objective. 

During inference, we obtain the importance score of documents in the validation set and test set based on Eq. \ref{eq::doc_rank}, and then reorder documents in descending order of their importance scores to create the meta-document. 

\noindent \textbf{Unsupervised Approach.} 
We hypothesize that the importance of a document is related to its centrality. To test this hypothesis, we propose an unsupervised centrality-based document reordering approach. %
To compute the centrality of a document $d_i$, we first represent the topic of the input cluster, $T_i$, by concatenating the top-3 sentences of each document except $d_i$, and then calculate the centrality as $\text{ROUGE}(d_i, T_i)$. 
We choose top-3 sentences to represent the topic as it is a strong unsupervised summarization baseline.
We avoid sentences of $d_i$ to be included in $T_i$ to prevent the centrality of $d_i$ being dominated by its own sentences, leading to similar centrality scores for all documents.

Finally, we reorder the documents in descending order of their centrality scores and then concatenate them into a meta-document.

\subsection{Base Extractive Summarization Model} %
\label{sec::base_model}
Once the documents have been reordered and concatenated to form a meta-document, they are fed to a base summarization model. \textbf{For the supervised reordering approach}, we use PreSumm \cite{liu2019text}, a state-of-the-art SDS method. For training, the extractive oracle labels are obtained by incrementally adding sentences to the extracted summary until the ROUGE score between the extracted summary and the gold abstractive summary does not increase. Using an SDS-based model architecture also facilitates transferring knowledge from SDS datasets. For this, we first finetune the model on SDS datasets and then finetune it on our MDS dataset. 
\textbf{For the unsupervised reordering approach}, we use PacSum \cite{zheng2019sentence}, a BERT-based model to measure the centrality of each sentence in the meta-document and then select sentences accordingly. We refer to Appendix \ref{app:base_model} for details of both approaches.

\section{Experiments}
In this section, we evaluate our document reordering based summarization approach. \footnote{Code is available at \url{https://github.com/zhaochaocs/MDS-DR}}

\subsection{Dataset}
We conduct experiments on two MDS datasets: Multi-News and DUC-2004. For evaluation, we compare the extracted summary to the gold abstractive summary. Due to the small size of DUC-2004, we use it only for out-of-domain evaluation. We also use CNN DailyMail (CNNDM) \cite{nallapati2016abstractive}, a single-document news summarization dataset, {to pretrain the base summarization model. } More details can be found in Appendix \ref{app::training}.

\subsection{Setup}
We use BERT\textsubscript{BASE} as the encoder of both the document reordering model and {base} summarization model. 
{We experiment with training the summarization model from scratch and also initializing it with parameters learned by training on CNNDM.} %
The details can be found in Appendix \ref{app::training}.
During inference, we choose the top-$K$ sentences with the highest score to compose the final summary, where $K$ is selected based on the average length of summaries in the training set. We set $K=9$ and $7$ for Multi-News and DUC-2004, respectively.

We compare our approach with the following baselines: Lead-$N$, TextRank \cite{mihalcea2004textrank}, LexRank \cite{erkan2004lexrank}, HiBERT \cite{zhang-etal-2019-hibert}, MGSum-ext \cite{jin-etal-2020-multi}, HDSG \cite{wang2020heterogeneous}, and MatchSum \cite{zhong2020extractive}. 
Lead-$N$ concatenates the top-$N$ sentence of each document. We try $N=\{1,2,3\}$ and report the best performance.
Following these approaches, we evaluate the extractive summaries using ROUGE $F_1$ score.\footnote{We use pyrouge (\url{https://github.com/bheinzerling/pyrouge})}

We evaluate the document reordering model by comparing the predicted document order with the oracle order via Kendall’s Tau ($\tau$) and Perfect Match Ratio (PMR), two common metrics for ranking tasks \cite{basu-roy-chowdhury-etal-2021-everything}. 
We compare our approach with a random baseline and a length-based baseline that rearranges documents in decreasing order of their lengths.\footnote{We do not include advanced baselines as performance of the document reordering is not the main focus of our work.} %

\subsection{Results}

\begin{table}[t!]
\input{tables/results}

 \caption{Summarization results evaluated on Multi-News by ROUGE 1 (R1), ROUGE 2 (R2), and ROUGE L (RL). Our best results (in \textbf{bold}) show %
 statistically significant difference with baselines (using paired bootstrap resampling, $p<0.05$ \cite{koehn2006manual}).}\vspace{-6pt}
 \label{tab:results}
\end{table}

\noindent \textbf{Automatic Evaluation} 
Table \ref{tab:results} shows results on Multi-News using either supervised or unsupervised document reordering  %
approach. 

We first investigate the utility of transferring knowledge from SDS. For this, we compare the PreSumm models trained from scratch (PreSumm w/o CNNDM) or %
{initialized using} %
CNNDM (PreSumm w/ CNNDM). Results show that PreSumm w/ CNNDM performs better than PreSumm w/o CNNDM (46.25 vs. 46.05 on ROUGE-1), indicating that the knowledge from SDS can be transferred to MDS by continual training. 
We then test the performance of our supervised document reordering (DR$_{\text{sup}}$) approach. Using document reordering, our approach, PreSumm + DR$_{\text{sup}}$, significantly outperforms the vanilla PreSumm on all ROUGE scores with or without CNNDM (46.57 vs. 46.25, 46.34 vs. 46.05 on ROUGE-1). Our unsupervised document reordering approach (DR$_{\text{unsup}}$) significantly outperforms PacSum on all ROUGE scores. 
These improvements demonstrate that document reordering is an effective way to leverage existing strong models for summarization. 

Our best approach, PreSumm + DR$_{\text{sup}}$, also significantly outperforms all of the baselines on all ROUGE scores. The performance gain is not entirely from the CNNDM, since our approach without CNNDM also achieves substantial improvements compared with all baselines. Similarly, the unsupervised approach, PacSum + DR$_{\text{unsup}}$, also outperforms the unsupervised baselines.
These improvements demonstrate that document reordering helps in multi-document summarization.

\noindent \textbf{Human Evaluation} We also conduct a human evaluation to better assess the performance of each system. We randomly select 100 test instances and evaluate the quality of a summary according to Informativeness, Conciseness, and Usefulness as in \citet{iskender2021reliability}. We conduct a pairwise comparison of PreSumm+DR$_{\text{sup}}$ (the best model) with PreSumm and MatchSum, two strongest neural baselines, as well as LEAD, the best unsupervised baseline. For each test instance, we obtain the output summary from our model and one of the baselines, and then ask three workers on Amazon Mechanical Turk to compare the two summaries according to the three measures listed above. More details can be found in Appendix \ref{app:human_eval}.

The results are shown in Table \ref{tab:res_human_eval}. Negative scores indicate worse performance compared with PreSumm+DR$_{\text{sup}}$. The results show that our approach can generate more informative, concise, and useful summaries compared to baselines, which is consistent with the automatic results.

\begin{table}[t!]
\setlength{\tabcolsep}{0.9em} %
\input{tables/human_eval}
 \caption{Results of human evaluation by comparing three baselines with PreSumm+DR$_{\text{sup}}$. A positive score means the baseline is better than ours and vise versa.}
 \label{tab:res_human_eval}
\end{table}

\begin{table}[t!]
\setlength{\tabcolsep}{1em} %
\input{tables/results_duc}

 \caption{Out-of-domain summarization results evaluated on DUC 2004 using the model trained on Multi-News. Our approach (last row) outperforms baselines. }
 \label{tab:results_duc}
\end{table}

\noindent \textbf{Out-of-domain Evaluation}
We further evaluate the performance of our approach in an out-of-domain setting. We compare our best approach with Lead-1, TextRank, MatchSum, and PreSumm. All models except Lead-1 and TextRank were trained on Multi-news and evaluated on the DUC 2004 dataset via Rouge $F_1$ scores. As shown in Table \ref{tab:results_duc}, our approach (last row of the table)
achieves consistently better performance than the baselines, indicating that our approach can effectively transfer to new unseen domains.

\begin{figure}[t]
    \centering
    \includegraphics[width=0.9\linewidth]{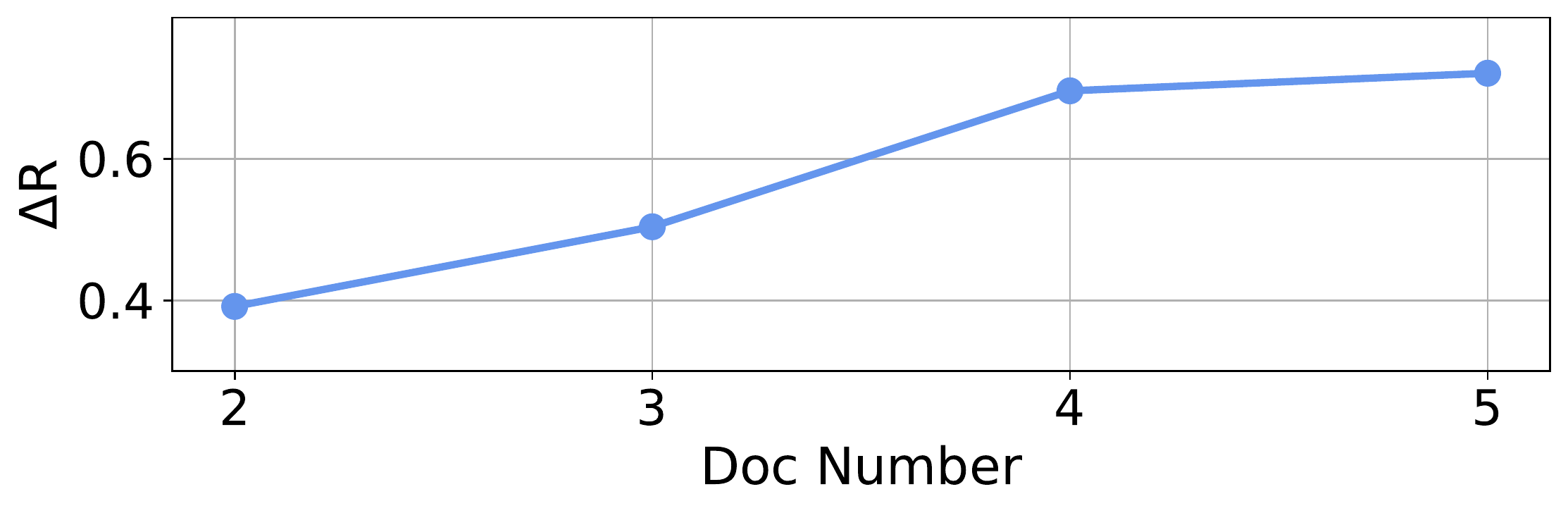}
    \caption{%
    Performance gain of summarization w.r.t. the number of input documents. We don't include instances with 6 or more documents since the number of such instances is small. Our approach results in more performance gain for longer inputs.} %
    \label{fig:doc_num}
\end{figure}

\begin{figure*}[!htb]
\minipage{0.27\textwidth}
  \includegraphics[width=\linewidth]{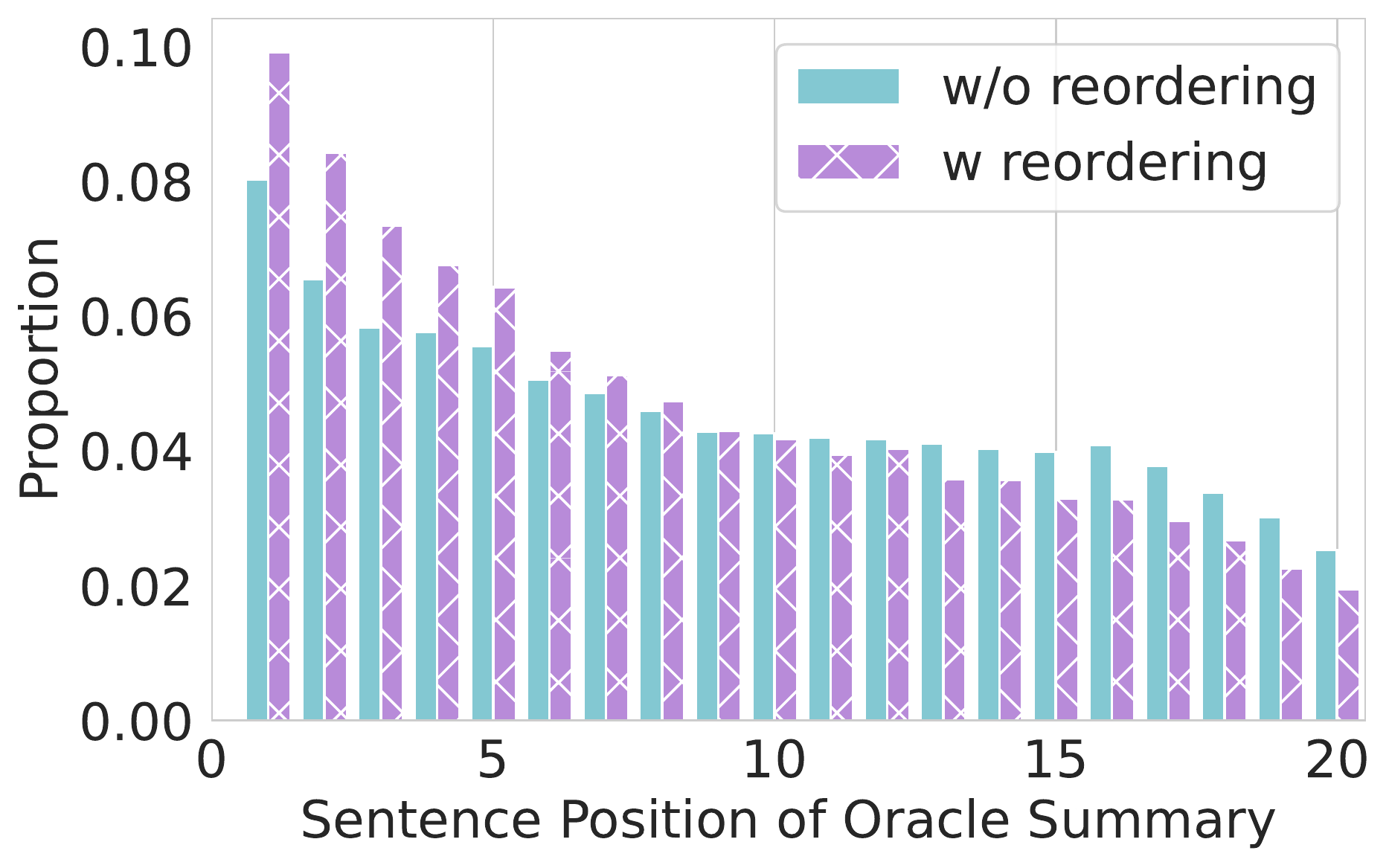}
\endminipage\hfill
\minipage{0.27\textwidth}
  \includegraphics[width=\linewidth]{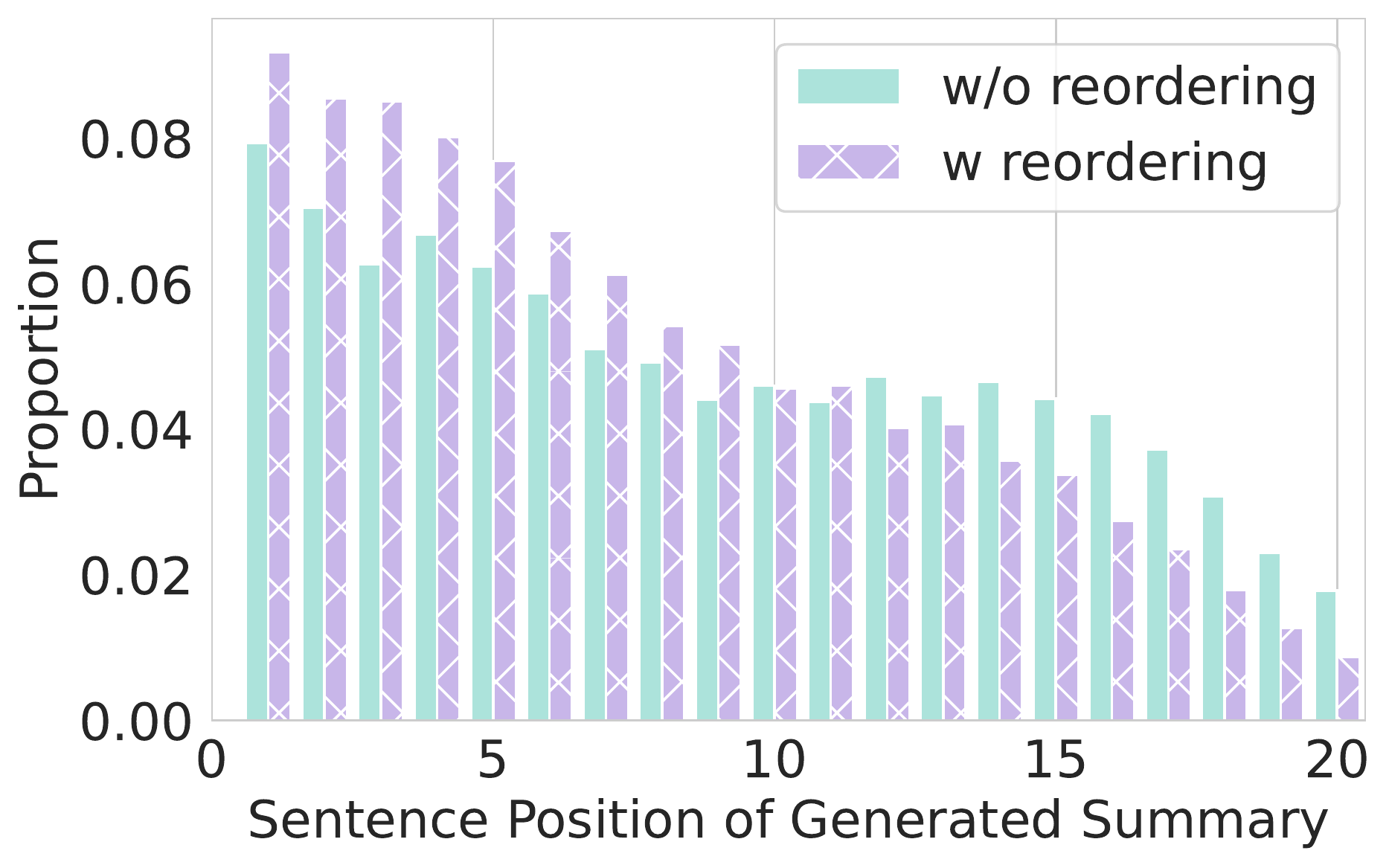}
\endminipage\hfill
\minipage{0.27\textwidth}%
  \includegraphics[width=\linewidth]{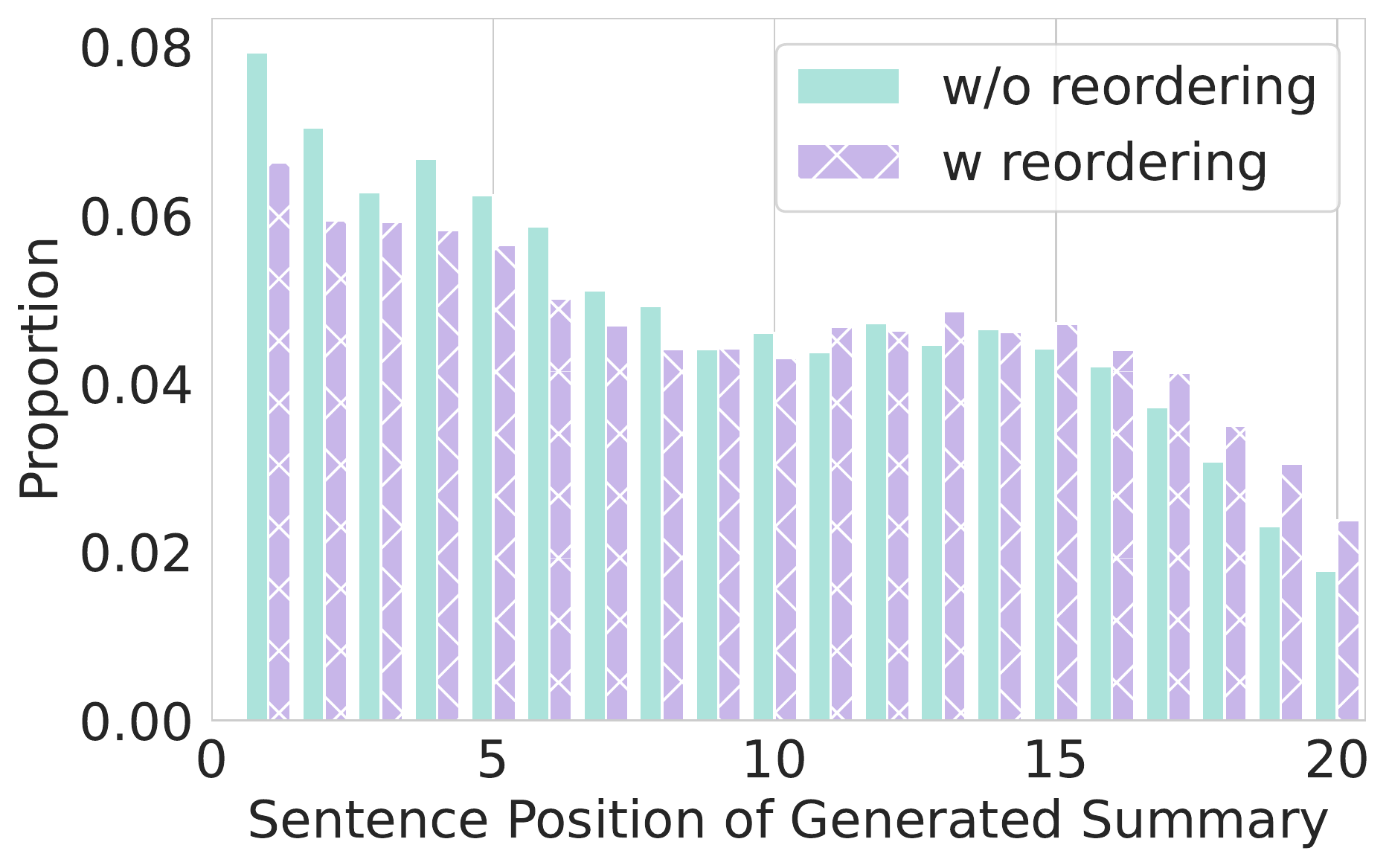}
\endminipage
\caption{(a) The distribution of \textit{oracle} extractive summaries according to their sentence positions in the meta-document with and without document reordering. (b) The distribution of \textit{generated} extractive summaries according to their sentence positions in the meta-document with and without document reordering. (c) The distribution of \textit{generated} extractive summaries according to their sentence positions in the original, unordered meta-document.}
\label{fig::analysis}
\end{figure*}

\subsection{Document-wise Analysis}
\begin{table}[t!]
\input{tables/results_ranking}
 \caption{Reordering methods evaluated on Multi-News. Our approaches, PreSumm + DR\textsubscript{sup} and PreSumm + DR\textsubscript{unsup} outperform the baselines.}
 \label{tab:results_rank}
\end{table}

In this section, we first compare our two %
document reordering approaches using 
ranking measures ($\tau$ and PMR) and ROUGE scores of the extracted summaries. Table \ref{tab:results_rank} shows the results. 
Our supervised ranking method (DR$_{\text{sup}}$) outperforms the unsupervised method (DR$_{\text{unsup}}$), demonstrating that the oracle importance score of the document is an effective supervision signal for document reordering. 
DR$_{\text{unsup}}$ achieves higher scores than baselines. It supports our hypothesis that the importance of documents is related to their centrality to the topic.

{We further analyze the impact of instance length (number of documents in the instance) on the model performance. In Figure \ref{fig:doc_num}, we group the test instances of Multi-News based on their lengths, and show the gain in summarization performance obtained from supervised reordering (measured using the ROUGE-1 difference $\Delta R$ between the models with and without document reordering). The figure shows that in general, $\Delta R$ increases as the instance length increases, indicating that instances with more documents benefit more from our reordering approach.} %

\subsection{Summary-wise Analysis}

The underlying assumption behind our document reordering approach is that extractive summarization models tend to select sentences from the beginning of the document. By reordering the important documents to the front of the meta-document, our approach  makes the salient content easier to learn. In this section, we investigate if this is indeed what is happening by analyzing the distribution of the oracle and the generated summary sentences in the meta-document. We conduct three experiments.

\textbf{Experiment 1: } 
We first investigate how reordering is changing the placement of important sentences. We represent important sentences as those in the oracle summaries, which is obtained by following the procedure described in Section \ref{sec::base_model}.
Figure \ref{fig::analysis}(a) shows the distribution of oracle summary-sentences at various positions of the input meta-document when it is reordered (purple shaded bars) and when it is not reordered (blue solid bars). The $x$-axis shows the sentence positions in the input meta-document and the $y$-axis shows the fraction of sentences from the oracle summary that were at that position in the meta-document. Comparing the purple and blue bars in the left area, more oracle summary's sentences were located at the beginning of the reordered input meta-document compared with the unordered input meta-document. This indicates that reordering helps in placing the important sentences in the beginning of the input meta-document.

\textbf{Experiment 2: } 
We next investigate if the summarization model favors certain sentence positions. Figure \ref{fig::analysis}(b) shows the distribution of (generated) summary-sentences with respect to various positions of the input meta-document for PreSumm+DR (w/ reordering, purple shaded bars) and PreSumm (w/o reordering, blue solid bars). Like Figure \ref{fig::analysis}(a), the $x$-axis shows the sentence positions in the input meta-document, but the $y$-axis shows the fraction of sentences from the \textit{generated} summary that are at that position in the meta-document. The bars on the left are, in general, higher than the bars on the right. This indicates that PreSumm tends to pick sentences appearing at the beginning of the input meta-document to create summaries.

\textbf{Experiment 3: } 
Finally, we want to investigate if the reordering can help the model select salient content that was originally scattered across the input. Figure \ref{fig::analysis}(c) shows the distribution of (generated) summary-sentences with respect to various positions of the \textit{original} unordered meta-document for PreSumm+DR (w/ reordering, purple shaded bars) and PreSumm (w/o reordering, blue solid bars). The $x$-axis shows the sentence positions in the \textit{original} meta-document and the $y$-axis shows the fraction of sentences from the generated summary that were at that position in the meta-document. We see that compared with the blue bars, the purple bars have a more uniform distribution. This indicates that the reordering based model has a greater tendency to pick sentences that were located at unfavorable positions (towards the end) in the original meta-document. The reordering helps in moving these sentences to the front, and then the summarization models pick them for generating the summary.

Overall, from these experiments, we can conclude that since the base summarization model pays more attention to the beginning of the input (Experiment 2), by moving important content towards the beginning of the input (Experiment 1), the reordering method helps the summarization model also focus on information that was scattered across the original unordered input (Experiment 3). %
We also provide a qualitative analysis in Appendix \ref{app:example} to show how the document reordering helps the model generate better summaries.

\section{Conclusion}
In this work, we propose a document reordering based approach for multi-document news summarization. We %
rearrange the documents according to their relative importance while concatenating them into a meta-document and then apply a summarization model. %
Our {simple yet effective} %
approach outperforms the baselines on two multi-document summarization datasets,
demonstrating that document reordering is a promising direction for %
multi-document news summarization. A next step, which we leave for future work, is to explore the scalability of such approaches on large document clusters.

\section*{Acknowledgements}
This work was supported in part by NSF grant IIS2112635.
We thank anonymous reviewers for their thoughtful and constructive reviews.

\section*{Ethical Considerations}
We do not foresee any ethical concerns from the technology presented in this work. We used publicly available datasets designed for summarization, and do not annotate any data manually. The datasets used is in English language.

\bibliography{anthology,custom}
\bibliographystyle{acl_natbib}

\appendix

\newpage

\begin{table*}[t]
    \centering
    \small
    \renewcommand{\arraystretch}{1.4}
    \input{tables/summ_example}
    \caption{Sample summaries generated by our method and the baselines. MatchSum and PreSumm receives the documents as the original order, making them focus more on the top two documents. Our method first rearrange the documents as the order of $\{3,4,2,1\}$ and then create the summary. We highlight the contents of the generated summaries which are relevant to the referenced summary. }
    \label{tab::example}
\end{table*}

\section{Base Summarization Models}
\label{app:base_model}
PreSumm \cite{liu2019text} is the supervised base summarization model. It uses BERT as the encoder to get the sentence representations, 
and a linear transformation with a Sigmoid as the decoder to get the probability of selecting a sentence. The loss function is the averaged cross-entropy between the predicted probability and the oracle 
$\{0,1\}$ label of each sentence. 
When applying this model to MDS, we insert a null sentence (``\texttt{[CLS]} \texttt{[SEP]}'') between consecutive (reordered) documents in the meta-document as the document delimiter. It helps the model to identify document boundaries and build inter-document relationships.

PacSum \cite{zheng2019sentence} is the unsupervised base summarization model. It uses sentence centrality to identify salient sentences. 
Different from other centrality-based methods, PacSum builds directed graph to explicitly model the order of sentences. Therefore PacSum can benefit from a meta-document where the salient documents are rearranged to the front.  When applying it to MDS, we build the graph for the meta-document and calculate the centrality of each sentence accordingly.

\section{Training Details}
\label{app::training}
We conduct experiments on Multi-News and DUC-2004. 
Multi-News is the largest multi-document summarization dataset in the news domain. It contains 44,972/5,622/5,622 instances for training/validation/test.
Each instance contains a set of news articles and an abstractive summary. The number of articles varies between 2 and 10. For evaluation, we compare the extracted summary to the gold abstractive summary. DUC-2004 contains 50 instances. Each instance has 10 documents and their abstractive summaries. Due to its small size, we use this dataset for out-of-domain evaluation only. 
We also use CNN DailyMail (CNNDM) to pretrain the base summarization model. It contains around 300K news articles and corresponding summaries from CNN and the Daily Mail.

We list the training details as follows. The training loss is optimized using Adam \cite{kingma2014adam} with a learning rate of $2 \times 10^{-3}$ and 10,000 training steps. We apply the warmup \cite{goyal2017accurate} on the first 2,000 steps and the early stopping based on the ROUGE-1 score on the development set. The batch size is set as 6,000 tokens. Our model was trained on a single Quadro RTX 5000 GPU in 2 hours.

\section{Human Evaluation Details}

\label{app:human_eval}
We randomly select 100 test instances to evaluate the performance of each system. The three measures we used are 1)  Informativeness: whether or not the summary reflects the salient information of the reference summary; 2) Conciseness: whether or not the summary contains no redundant words or repeated information; and 3) Usefulness: whether or not the summary helps the reader catch the main idea of the news. Human judges were paid at a wage rate of \$8 per hour, which is higher than the local minimum wage rate.

The pairwise scores of those measures are calculated as follows. When comparing a certain baseline approach to our model, we report the percentage of summaries created by the baseline that were judged to be better/worse/same than those of our model, yielding a score ranging from -1 (unanimously worse) to 1 (unanimously better). For example, when evaluating the informativeness scores, Lead performs better/worse/same than our model for 36\%/56\%/8\% of the instances, yielding a pairwise score as 0.36-0.56=-0.20.

\section{Qualitative Analysis}
\label{app:example}
Table \ref{tab::example} shows an example with 4 source documents listed in the original order. The main event of this example is about a child abduction case, where source 3 and 4 provide more direct and detailed information compared with source 1 and 2.

We show the summaries generated by MatchSum, PreSumm, and our system, as well as the reference summary. MatchSum and PreSumm receive the documents in the original order, making them focus more on the top two documents. Our method first rearranges the documents as the order of $\{3,4,2,1\}$ and then creates the summary based on the new re-ordered documents. With the help of the document reordering, our summary better captures the main event from the latter source documents (source 3 and source 4).

\end{document}

%% file: tables/results.tex
\setlength\tabcolsep{3pt} %
\small
\centering
 \begin{tabular}{l c c c} 
 \toprule[1pt]
 \textsc{Model} & \textsc{R1} & \textsc{R2} & \textsc{RL} \TBstrut\\ 
 \midrule[1pt]
 Lead \cite{fabbri2019multi} & 43.08 & 14.27 & 38.97\\
 LexRank \cite{erkan2004lexrank} & 41.77 & 13.81 & 37.87 \\
 TextRank \cite{mihalcea2004textrank} & 41.95 & 13.86 & 38.07 \\
 HiBERT \cite{zhang-etal-2019-hibert} & 43.86 & 14.62 & - \\
 MGSum-ext \cite{jin-etal-2020-multi} & 44.75 & 15.75 & - \\
 \text{HDSG} \cite{wang2020heterogeneous} & 46.05& 16.35& 42.08 \\
 MatchSum \cite{zhong2020extractive} & 46.20 & 16.51 & 41.89\\
 \midrule[1pt]
 \rowcolor{Gray!93} \textit{Unsupervised} & & & \Tstrut\\
 \textsc{PacSum} & 43.02 & 14.03 & 39.02 \\ 
 \textsc{PacSum} + DR\textsubscript{unsup} \text{(Ours)} & 43.57 & 14.41 & 39.52 \Bstrut\\
 \midrule[1pt]
 \rowcolor{Gray!93} \textit{Supervised, w/o finetune on CNNDM} & & & \Tstrut\\
 \textsc{PreSumm} & 46.05 & 16.56 & 41.91 \\ 
 \textsc{PreSumm} + DR\textsubscript{sup} \text{(Ours)} & 46.34 & 16.88 & 42.20 \Bstrut\\
 \midrule[1pt]
 \rowcolor{Gray!93} \textit{Supervised, w/ finetune on CNNDM} & & & \Tstrut\\
 \textsc{PreSumm} & 46.25 & 16.75 & 42.11 \\ 
 \textsc{PreSumm} + DR\textsubscript{sup} \text{(Ours)} & \textbf{46.57} & \textbf{17.10} & \textbf{42.44} \Bstrut\\
 \midrule[1pt]
 Oracle  & 49.06 & 21.54 & 44.27 \TBstrut\\
 \bottomrule[1pt]
 \end{tabular}

%% file: tables/human_eval.tex
\small
\centering
\begin{tabular}{l|ccc}
\toprule
Model & Informative & Concise & Useful \\\midrule
LEAD & -0.20 & -0.14 & -0.17 \\
MatchSum & -0.12 & -0.05 & -0.08 \\
PreSumm & -0.06 & 0.03 & -0.07 \\\bottomrule
\end{tabular}

%% file: tables/results_duc.tex
\small
\centering
 \begin{tabular}{l | c c c} 
 \toprule
 \textsc{Model} & \textsc{R1} & \textsc{R2} & \textsc{RL} \TBstrut\\ 
 \midrule
  Lead-1 & 33.86 & 7.51 & 29.64 \\
 TextRank & 33.09 & 7.49 & 29.25 \\
 MatchSum  & 33.84 & 7.44 & 30.07 \\
 PreSumm & 34.42 & 7.95 & 30.34 \\  
 PreSumm + DR\textsubscript{sup} & 34.62 & 8.22 & 30.54 \\ 
 \bottomrule
 \end{tabular}

%% file: tables/results_ranking.tex
\small
\centering
 \begin{tabular}{l |c c |c c c} 
 \toprule
 \multirow{2}{*}{\textsc{Model}} & \multicolumn{2}{c|}{Reordering} & \multicolumn{3}{c}{Summarization}  \\
    & $\tau$ & PMR & R1 & R2 & RL \TBstrut\\
 \midrule
 Random & -0.005 & 31.8 & 46.25 & 16.75 & 42.11 \\
 Length & 0.189 & 43.2 & 46.30 & 16.73 & 42.15 \\
 DR\textsubscript{unsup} & 0.236 & 46.4 & 46.41 & 16.94 & 42.26 \\
 DR\textsubscript{sup} & 0.325 & 51.7 & 46.57 & 17.10 & 42.44 \\
 \bottomrule
 \end{tabular}

%% file: tables/summ_example.tex
\begin{tabular}{p{0.99\linewidth}}
    \toprule
         \textbf{Source 1:} these items are among those purchased by gary simpson , prior to taking 9-year-old carlie trent from her school in rogersville , tn on may 4th ... share this : twitter facebook linkedin google email like this : like loading ... \\
         
         \textbf{Source 2:} by hayes hickman of the knoxville news sentinel two knoxville banking executives are offering a \$ 10,000 reward for information leading to the return of missing 9-year-old carlie marie trent, who was abducted a week ago by her uncle in hawkins county . matt daniels , president and chief executive officer of apex bank , said he and his business partner , 21st mortgage president tim williams , felt compelled to get involved as the search continues ... “this is agonizing , ” daniels said wednesday ." “it ’ s not a 24-hour amber alert. ” daniels said carlie reminds him of his two daughters ... \\

      \textbf{Source 3:} it ' s been a week since carlie trent was in school . 
      on wednesday , tennessee bureau of investigation asked the public to trust them that carlie trent really is in danger . josh devine , a spokesperson for tbi , said he has heard some disturbing rumors on social media that if simpson was trying to protect carlie trent . he told wate he has seen tweets that asked " if simpson didn ' t harm carlie when he had custody , why would he do it now ." " i don ' t think he ' s trying to protect her ." this was not an innocent camping trip , this was a crime . \\
      
      \textbf{Source 4:} the mother of missing tennessee 9-year old carlie trent says she \" always had a bad feeling \" about gary simpson , the 57-year-old man who allegedly kidnapped her daughter last week on may 4 . simpson , carlie ' s uncle by marriage who , along with his wife , had once had custody of carlie , picked her up from her rogersville elementary school . simpson and carlie did not return home later that afternoon and the following day an amber alert was issued in tennessee ... \\

      \midrule
         \textbf{MatchSum:} by hayes hickman of the knoxville news sentinel two knoxville banking executives are offering a \$ 10,000 reward for information leading to the return of \textbf{\textcolor{violet}{missing 9-year-old carlie marie trent , who was abducted a week ago by her uncle in hawkins county }}. matt daniels , president and chief executive officer of apex bank , said he and his business partner , 21st mortgage president tim williams , felt compelled to get involved as the search continues ... “ this is agonizing , ” daniels said wednesday . “ it ’ s not a 24-hour amber alert . ” daniels said carlie reminds him of his two daughters ...  \\
         \textbf{PreSumm:} by hayes hickman of the knoxville news sentinel two knoxville banking executives are offering a \$ 10,000 reward for information leading to the return of \textbf{\textcolor{violet}{missing 9-year-old carlie marie trent , who was abducted a week ago by her uncle in hawkins county .}} matt daniels , president and chief executive officer of apex bank , said he and his business partner , 21st mortgage president tim williams , felt compelled to get involved as the search continues ... mother of allegedly abducted 9-year-old carlie trent \textbf{\textcolor{violet}{' always had a bad feeling ' about suspect}} . these items are among those purchased by gary simpson , prior to \textbf{\textcolor{violet}{taking 9-year-old carlie trent from her school}} in rogersville , tn on may 4th ...  \\
         \textbf{Ours:} the mother of \textbf{\textcolor{violet}{missing tennessee 9-year old carlie trent}} says she \textbf{\textcolor{violet}{" always had a bad feeling " about gary simpson , the 57-year-old man}} who allegedly \textbf{\textcolor{violet}{kidnapped her daughter last week on may 4 , simpson, carlie ' s uncle ... picked her up from her rogersville elementary school . }} he told wate he has seen tweets that asked " if simpson didn ' t harm carlie when he had custody , why would he do it now . “it ’ s not a 24-hour amber alert. \textbf{\textcolor{violet}{this was not an innocent camping trip , this was a crime}} . \textbf{\textcolor{violet}{" i don ' t think he ' s trying to protect her ."}} simpson and carlie did not return home later that afternoon and the following day an amber alert was issued in tennessee ... \\\midrule
         
         \textbf{Reference:} – authorities are combing through more than 1,200 leads in a desperate search for \textbf{\textcolor{violet}{a 9-year-old girl they say was abducted by her uncle may 4}} , wate reports . according to the knoxville news sentinel , \textbf{\textcolor{violet}{57-year-old gary simpson picked carlie trent up from her tennessee school }} ... the tbi says \textbf{\textcolor{violet}{there have been rumors online that simpson is trying to protect carlie}} , but it says that couldn ' t be further from the truth . \textbf{\textcolor{violet}{this was not an innocent camping trip , this was a crime}} ...
         shannon trent , who hasn \' t had custody of carlie in two years , says \textbf{\textcolor{violet}{she " always had a bad feeling " about simpson}} ...\\\bottomrule
    \end{tabular}
    